\documentclass[10pt,twocolumn,letterpaper]{article}

\usepackage{cvpr}              

\usepackage[dvipsnames]{xcolor}

\definecolor{cvprblue}{rgb}{0.21,0.49,0.74}
\usepackage[pagebackref,breaklinks,colorlinks,citecolor=cvprblue]{hyperref}

\def\paperID{10767} 
\def\confName{CVPR}
\def\confYear{2024}

\title{Boosting3D: High-Fidelity Image-to-3D by Boosting 2D Diffusion Prior to 3D Prior with Progressive Learning}

\author{Kai Yu,~~~Jinlin Liu,~~~Mengyang Feng,~~~Miaomiao Cui,~~~Xuansong Xie\\
Alibaba Group\\
{\tt\small \{jinmao.yk, ljl191782, mengyang.fmy, miaomiao.cmm, xingtong.xxs\}@alibaba-inc.org}
}

\begin{document}

\twocolumn[{%
\renewcommand\twocolumn[1][]{#1}%
\maketitle
\begin{center}
    \centering
    \captionsetup{type=figure}
    \includegraphics[width=0.95\textwidth]{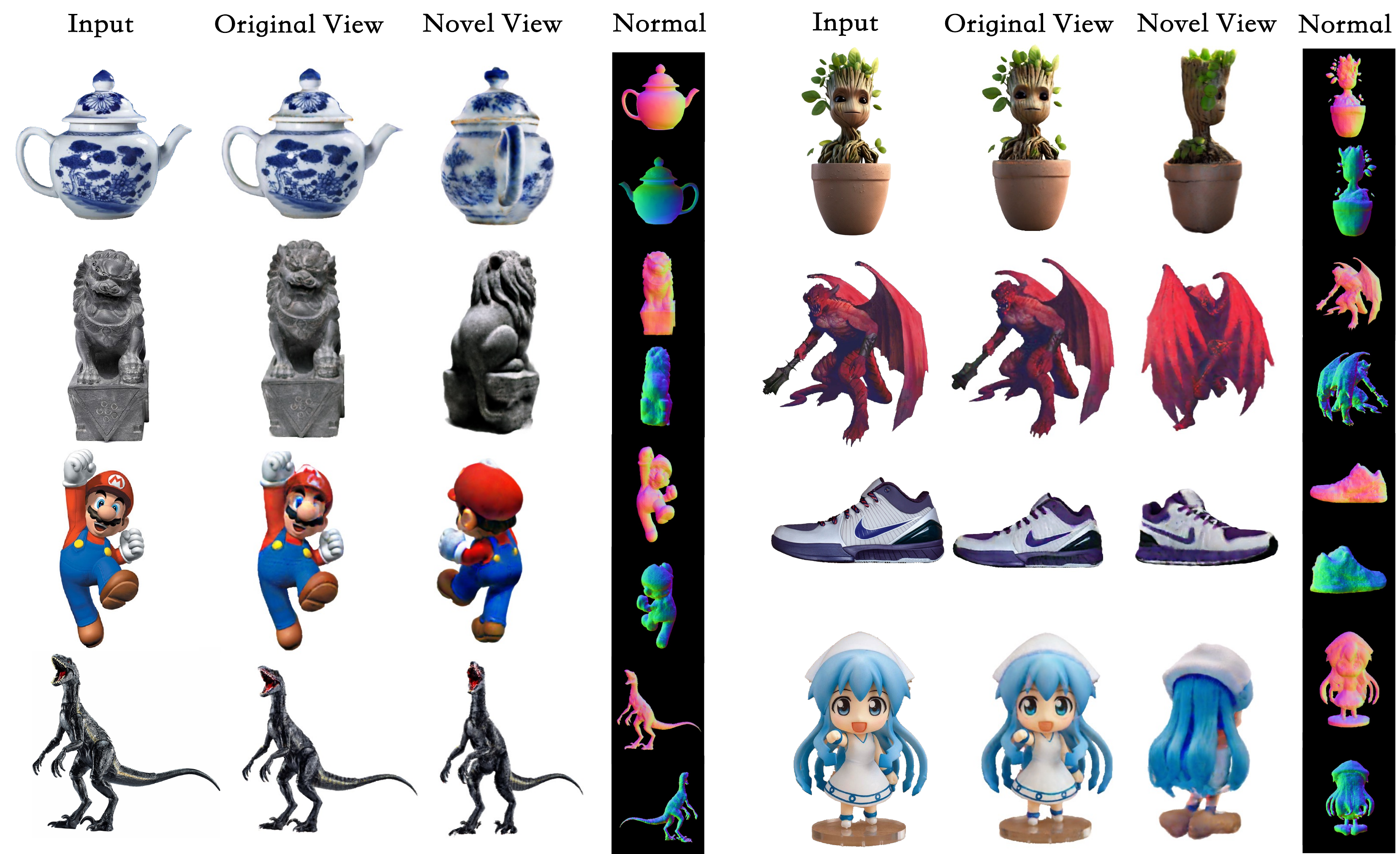}
    \captionof{figure}{The results of Boosting3D for the image-to-3d generation task. Our method can reconstruct reasonably detailed 3D mesh from a single image in different data domains.}
    \label{'resimg'}
\end{center}%
}]
\maketitle

\begin{abstract}
\vspace{-0.1cm}
We present Boosting3D, a multi-stage single image-to-3D generation method that can robustly generate reasonable 3D objects in different data domains. 
The point of this work is to solve the view consistency problem in single image-guided 3D generation by modeling a reasonable geometric structure. For this purpose, we propose to utilize better 3D prior to training the NeRF. 
More specifically, we train an object-level LoRA for the target object using original image and the rendering output of NeRF. And then we train the LoRA and NeRF using a progressive training strategy. The LoRA and NeRF will boost each other while training. After the progressive training, the LoRA learns the 3D information of the generated object and eventually turns to an object-level 3D prior. In the final stage, we extract the mesh from the trained NeRF and use the trained LoRA to optimize the structure and appearance of the mesh. The experiments demonstrate the effectiveness of the proposed method. Boosting3D learns object-specific 3D prior which is beyond the ability of pre-trained diffusion priors and achieves state-of-the-art performance in the single image-to-3d generation task. 
\end{abstract}

\section{Introduction}
\label{sec:intro}

Estimating 3D model from only one input image is challenging for the ambiguity and the complexity of real world objects. Many previous works \cite{zheng2019deephuman,kolotouros2019convolutional,saito2019pifu,saito2020pifuhd,xiu2023econ} focus on only some particular category, such as human, due to the promising application. 3D human dataset are collected first to train a 3D network. Whereas, these kinds of methods are not applicable to open-vocabulary image-to-3D task due to the lack of diverse 3D datasets. 

To solve the dataset problem, some previous works try to learn 3D structure from only 2D image collections \cite{skorokhodov20233d,sargent2023vq3d}. 2D image collections such as ImageNet \cite{deng2009imagenet} contain diverse images with different view angles. And thus, 3D structures can be learned from these 2D images. Recent amazing progress in diffusion models makes it possible to generate diverse images with only text prompt. 2D diffusion models are trained using billions of images LAION 5B \cite{schuhmann2022laion}, which contain object photos taken from different views. Well-trained 2D diffusion models can thus be used to learn 3D structures for open-vocabulary objects. 

Many recent works \cite{dreamgaussian,chen2023fantasia3d,wang2023prolificdreamer,poole2022dreamfusion} use 2D diffusion prior for text-to-3D generation. 3D representation networks such as NeRF or DMtet~\cite{dmtet} are trained using pretrained 2D diffusion models with SDS \cite{poole2022dreamfusion} or VDS loss \cite{wang2023prolificdreamer}. Furthermore, some recent works use diffusion prior to solve open-vocabulary image-to-3D task. Image-to-3D aims at estimating 3D structure given an input image. As the input image may be different from typical generated images from 2D diffusion model, it becomes more difficult to train than text-to-3D task. Zero-1-to-3 \cite{liu2023zero} trains a diffusion model using multiple views of images rendered from 3D dataset. This trained diffusion model using 3D dataset is more powerful than normal pretrained diffusion models in respect to 3D capability and is referred as 3D diffusion prior. Lots of recent methods \cite{tang2023make,melas2023realfusion,liu2023one,qian2023magic123,xu2023neurallift} manage to train a 3D representation network with only one given image using 2D diffusion prior or 3D diffusion prior.  

Although amazing progress has been achieved by recent methods, we notice that it may fail when the input image containing uncommon objects with asymmetry structure, such as objects from video games. These kinds of irregular object are beyond the ability of normal 2D diffusion prior and 3D diffusion prior. Because of this, we propose Boosting3D to boost normal 2D diffusion prior to 3D diffusion prior with progressive learning. 

First, we optimize a coarse NeRF using the pretrained diffusion models. Simutaneously, we train a LoRA for the specific input object. Next we train the LoRA and NeRF in a progressive way. The LoRA and NeRF will boost each other while training.  After this step, we obtain a refined NeRF and a well trained LoRA with object-level 3D prior. Finally, we extract a coarse surface mesh from the trained NeRF and finetune both surface geometry and appearance using the trained LoRA. Our method is able to obtain high-quality and stable 3D object from one input image as shown in Fig.\ref{'resimg'}. In summary, we make the following three main contributions:

\begin{itemize}
    \item We present Boosting3D, a novel image-to-3D pipeline that uses three-stage optimization process, i.e. coarse NeRF, fine NeRF and mesh refinement, to generate a high-quality textured mesh.
    \item We propose a novel 3D mesh optimization method that can explicitly optimize 3D model representation and texture using T2I model. The proposed method outperforms explicit 3D representation method DMtet in terms of mesh and texture quality.
    \item We boost 2D diffusion prior to 3D prior in a bootstrap way by training object-level LoRA . Our method achieves state-of-the-art results in 3D reconstruction of single objects for both real-world photos and synthetic images. 
\end{itemize}


\section{Related work}
\label{sec:related}

\subsection{Diffusion models}

Recently, large-scale diffusion models have shown great performance in text-to-image synthesis \cite{ho2020diffusion}, which provides an opportunity to utilize it for zero-shot text-to-3D generation \cite{wang2023prolificdreamer,poole2022dreamfusion,lin2023magic3d,guo2023threestudio}. 
LoRA\cite{hu2021lora} propose to use the low rank matrix to learn the generation information of a category or object, reducing the amount of trained parameters. Dreambooth\cite{dreambooth} propose a training method that uses a fixed prompt and a small number of samples to finetune the whole model. Both methods enable learning the specific object-level information at a low cost.

To acquire different views of  the input image,  Zero-1-to-3 \cite{liu2023zero} and syncdreamer\cite{liu2023syncdreamer} train a diffusion model using multiple views of images rendered from 3D dataset. The trained diffusion model can then be used to generated multiple views of the given image. For the capability of generating multiple views,  this diffusion model is treated as 3D diffusion prior


\subsection{Text-to-3D generation}

The goal of text-to-3D task is to generate a 3D model that is consistent with the semantics of the input prompt. Dreamfusion\cite{poole2022dreamfusion} proposes score decomposition sampling (SDS) loss to generate 3D models, which aims to minimize the distribution difference between NeRF\cite{nerf} rendering and pre-trained text-to-image (T2I) models. Latent-nerf\cite{metzer2023latent} improves the performance of 3D generation by optimizing NeRF in latent space. In addition to generating 3D objects, SDS loss can also work in scene generation\cite{zhuang2023dreameditor}. Some works\cite{chen2023fantasia3d,dreamgaussian,lin2023magic3d,ts2023textmesh} use other 3D representation methods but also used SDS loss for optimization. Prolificdreamer\cite{wang2023prolificdreamer} propose variable score decomposition (VSD) loss, which can generate high-quality and high-fidelity results. 
Text-to-3d method\cite{raj2023dreambooth3d,jun2023shape,gao2022get3d,cao2023texfusion,richardson2023texture} uses prompt to control views when generating 3D views, which may lead to multi-face problem. Dreamtime\cite{huang2023dreamtime} controls the change of noise sampling level during the generation process to mitigate multi-face problem. As text prompt is not accurate enough to describe 3D model, some other methods using image guidance to generate 3D model.


\subsection{Image-to-3D generation}

The image-to-3d task can be regarded as a task of 3D reconstruction from a single image\cite{yu2021pixelnerf,duggal2022topologically}. Previous single image reconstruction works focus on fixed class reconstruction tasks\cite{saito2019pifu,saito2020pifuhd,xiu2023econ}, which often require a large-scale 3D training data. The difficulty of obtaining 3D data makes it not applicable to open-vocabulary objects. The text-to-image model trained with large amount of images contains 3D related information, which is the key of single image zero-shot reconstruction. Make-it-3D\cite{tang2023make} introduces SDS loss into image-to-3d task, and uses pre-trained diffusion model and clip\cite{clippaper} model to complete 3D generation. Models such as zero123\cite{liu2023zero} and syncdreamer\cite{liu2023syncdreamer} can directly generate multi-view of the input image for multi-view reconstruction. Limited by the training data, the multi-view generated can not guarantee the complete 3D consistency for open-vocabulary inputs. Magic123\cite{qian2023magic123} uses zero123 and pre-trained diffusion model as priors, which can achieve high-quality single image guided 3D generation. Dreamgaussion\cite{dreamgaussian} and one-2-3-45\cite{liu2023one} uses the new 3D representation combined with diffusion model to achieve rapid 3D generation. 

The above methods\cite{tang2023make,qian2023magic123,dreamgaussian} are optimized by SDS loss using pre-trained diffusion priors.  We notice that these methods may fail when the input image containing uncommon objects with asymmetry structure, such as objects from video games. These kinds of irregular object are beyond the ability of normal 2D diffusion prior and 3D diffusion prior.  To solve this, we introduce object-specific LoRA to boost 2D diffusion prior to 3D prior. Moreover, we optimize the texture and structure of the extracted mesh using the trained LoRA, generating high-quality 3d model.

 


\section{Pipeline}

\begin{figure*}
  \centering
  \includegraphics[scale=0.6]{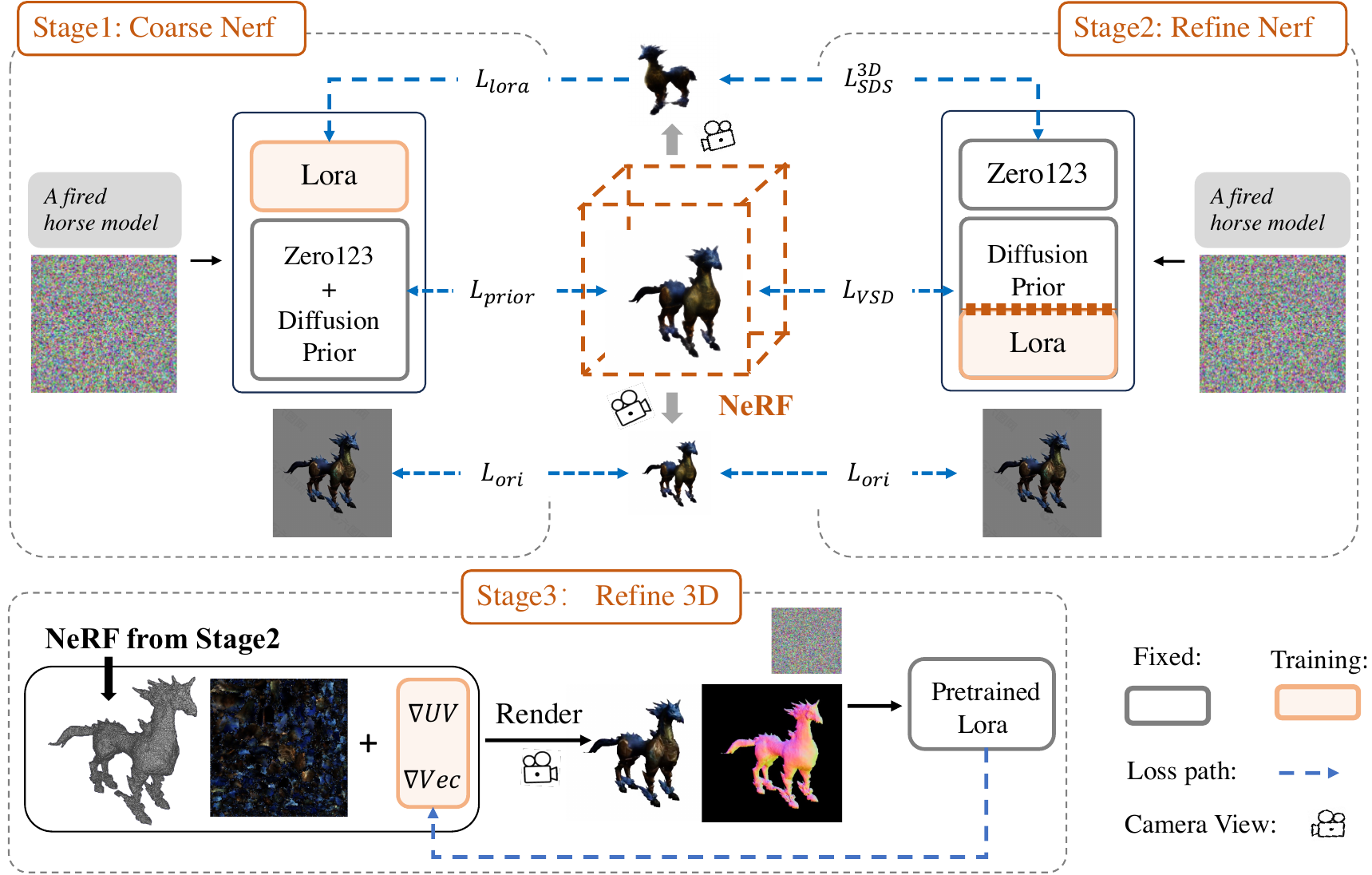}
  \caption{The pipeline of Boosting3D. Boosting3D is a three-stage framework for high quality 3D generation from a reference image. In stage 1, we optimized a course NeRF and a object-level LoRA. In stage 2, we refined the NeRF using the pre-trained model and the LoRA trained in stage 1. In stage 3, we extracted the 3D mesh from the trained NeRF and refined the 3D model using the pre-trained LoRA.}
  \label{fig1}
\end{figure*}

In this section, we introduce Boosting3D, a three-stage pipeline for Image-to-3D task as illustrated in Fig.\ref{fig1} and present preliminaries on score distillation sampling, variational score distillation and multi-views generation (Section \ref{sec:pre}). Firstly, we optimze a NeRF using pretrained model, and train a LoRA initialize the object-level information (Section \ref{sec:s1}). Next we train the LoRA and NeRF in a progressive way. The LoRA and NeRF boost each other during training.  After this step, we obtain a refined NeRF and a well trained LoRA with object-level 3D prior. (Section \ref{sec:s2}). Finally, we extract a coarse surface mesh from trained NeRF and fine-tune both surface geometry and appearance using trained LoRA (Section \ref{sec:s3}).

\subsection{Preliminaries}
\label{sec:pre}

Many text-to-3D and image-to-3D methods use large-scale diffusion models as an optimization foundation. Dreamfusion\cite{poole2022dreamfusion} uses pretrained diffusion model $\epsilon_\phi$ to realize the conversion from text to 3D model, which proposes score distillation sampling (SDS) loss to use prompt $y$ to guide 3D model $\theta$ generation. SDS loss encourages the trained 3D model to sample image information from the pretrained diffusion models, so that the 3D rendering results $x$ are consistent with the diffusion models distribution mode. Specifically, the SDS loss computes the gradient:

\begin{equation}
\nabla_{\theta}L_{SDS}=\mathbb{E}_{t,\epsilon,p}\left[w_t\left(\epsilon_\phi\left(x_t^p;t,y\right)-\epsilon\right)\frac{{\partial}x^p}{\partial\theta}\right]
  \label{eq:sds}
\end{equation}

where $ \epsilon_\phi$(·) is the predicted noise by the 2D diffusion prior $\phi$, $x_t^p$ is the render image $x_t^p$ in view $p$ add noise at the noise level $t$, $w_t$ is a weight about $t$. SDS loss can realize the conversion of text to 3D, but suffers from over-saturation, low-diversity, and smoothing problems. ProlificDreamer\cite{wang2023prolificdreamer} proposed variational score distillation (VSD) loss to solve these problem, which can obtain more refined 3D representation and texture. Different from SDS in minimizing the image distribution, VSD uses LoRA to sample distribution in the pre-trained space, which can produce results with photorealistic rendering. The VSD loss computes the gradient:

\begin{equation}
\nabla_{\theta}L_{VSD}=\mathbb{E}_{t,\epsilon,p}\left[w_t\left(\epsilon_\phi\left(x_t;t,y\right)-\epsilon_{lora}\left(x_t^p;t,y,c\right)\right)\frac{{\partial}x^p}{\partial\theta}\right]
  \label{eq:vsd}
\end{equation}

where $ \epsilon_{lora} $ estimates the score of the rendered images using a LoRA (Low-rank adaptation) model.

In addition to the text-to-image model, there are also some models specially trained to generate multi-views. Such models contain more accurate 3D information of objects, such as Zero123XL \cite{liu2023zero} used in this paper. For Zero123XL, input an image $x^0$ and the viewing angle difference with the input image to generate an image corresponding to the viewing angle. For Zero123XL, the gradient of SDS loss can be changed to the following form;

\begin{equation}
\nabla_{\theta}L_{SDS}^{3D}=\mathbb{E}_{t,\epsilon,p}\left[w_t\left(\epsilon_\phi\left(x_t^p;t,x^0,{\Delta}p\right)-\epsilon\right)\frac{{\partial}x^p}{\partial\theta}\right]
  \label{eq:sds_3d}
\end{equation}

where ${\Delta}p$ is the camera pose difference between the current view $x^p$ and the input view $x^0$.

\subsection{Stage1: Coarse NeRF Generation}
\label{sec:s1}

In the first stage, we obtain a coarse NeRF model that can correspond to the objects in the input image. In the process of training the NeRF model, we divide the training views into two modes: the original view of input image, using the original image as supervision; the new views of the object, using pre-trained text-to-image model and pre-trained 3D priors (Zero123XL) as supervision.

For the original view of the input image $I_0$, we obtain image $I$ and corresponding mask $M$ through NeRF rendering. Here we use the original image to calculate $L1$ loss for $I$, use MSE loss to calculate the loss of the original image corresponding to mask $M_0$ and $M$, and add corresponding weights to the two losses to obtain Loss:



\begin{equation}
L_{ori}= {\lambda}_{rgb}{\parallel I_0 - I {\parallel}_1} + {\lambda}_{mask}{\parallel M_0 - M {\parallel}_2^2}
  \label{eq:ori}
\end{equation}

For new view of the object, we render the current image through NeRF to obtain the image $I_n$ and normal map $N_n$. We add noise to $I_n$ and then input it into the pre-trained 3D prior model and the pre-trained T2I model to obtain the SDS loss of both, and add the corresponding weights to the two losses. The gradient consists of Eq.\ref{eq:sds} and Eq.\ref{eq:sds_3d}:

\begin{equation}
\nabla_{\theta}L_{prior}= {\lambda}_{sds} \nabla_{\theta}L_{SDS} + {\lambda}_{3d}\nabla_{\theta}L_{SDS}^{3D}
  \label{eq:prior_s1}
\end{equation}

The model corresponding to NeRF at this stage will have a lot of noise, so we added 2D normal map smooth loss to make the overall NeRF smoother:

\begin{equation}
L_{normal}= {\lambda}_{normal}{\parallel N_n - {\delta}(N_n) {\parallel}_2^2}
  \label{eq:normal}
\end{equation}

where ${\delta}$(·) represents the result of moving the normal map by 1 pixels to random direction.

\begin{figure}[t]
  \centering
   \includegraphics[scale=0.5]{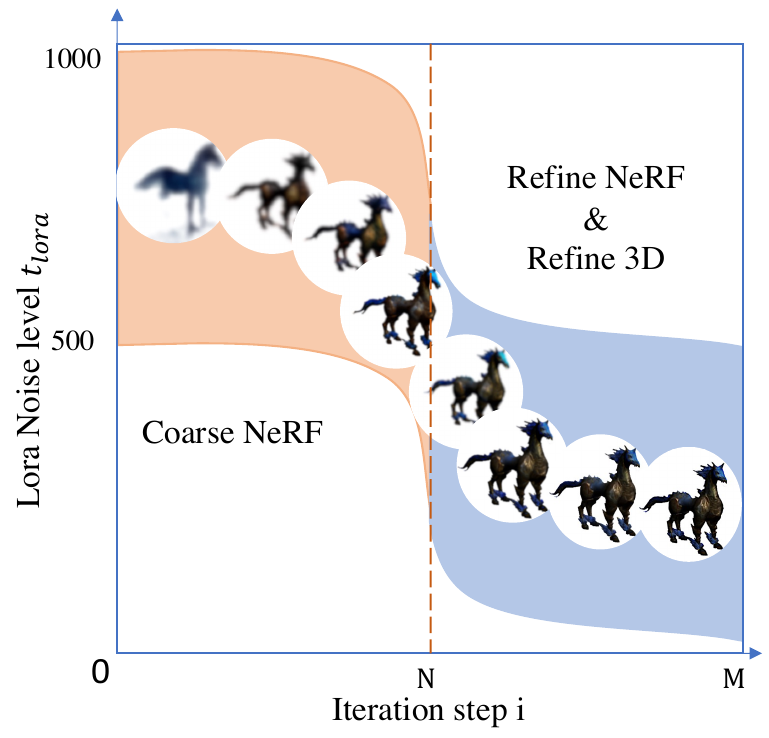}
   \caption{The proposed noise level for training. We use a higher noise level to train in stage 1 and use a lower noise level in stage 2$\&$3. N represents the training steps in stage 1, and M represents the training steps in total.}
   \label{fig2}
\end{figure}

In the first stage, we will train a LoRA in the process of training NeRF based on the original image and the render image of NeRF, which will use a higher noise level $t_{lora}$ when training LoRA, as shown in Fig.\ref{fig2}. 

\begin{equation}
L_{lora}=\parallel \epsilon_{lora}\left(x_t^p;t_{lora},y,c\right)-\epsilon
   {\parallel}_2
   \label{eq:lora}
\end{equation}

In practice, we parameterize $\epsilon_{lora}$ by a LoRA of the pretrained model $\epsilon_\phi$, and use camera parameter $c$ as the class embeddings. The LoRA will serve as the initialization of LoRA in the second stage.

Overall, the stage 1 is optimized by $L_{s1}$:


\begin{equation}
L_{s1}= L_{ori} + L_{prior} + L_{normal} + L_{lora}
  \label{eq:S1}
\end{equation}

In process of training, We alternately train NeRF using the original input image and the new view, while training LoRA using the rendering results of the NeRF. And we find that using a specific range of noise level can make the results more refined and fit the input image.

\begin{figure*}[t]
  \centering
   \includegraphics[scale=0.37]{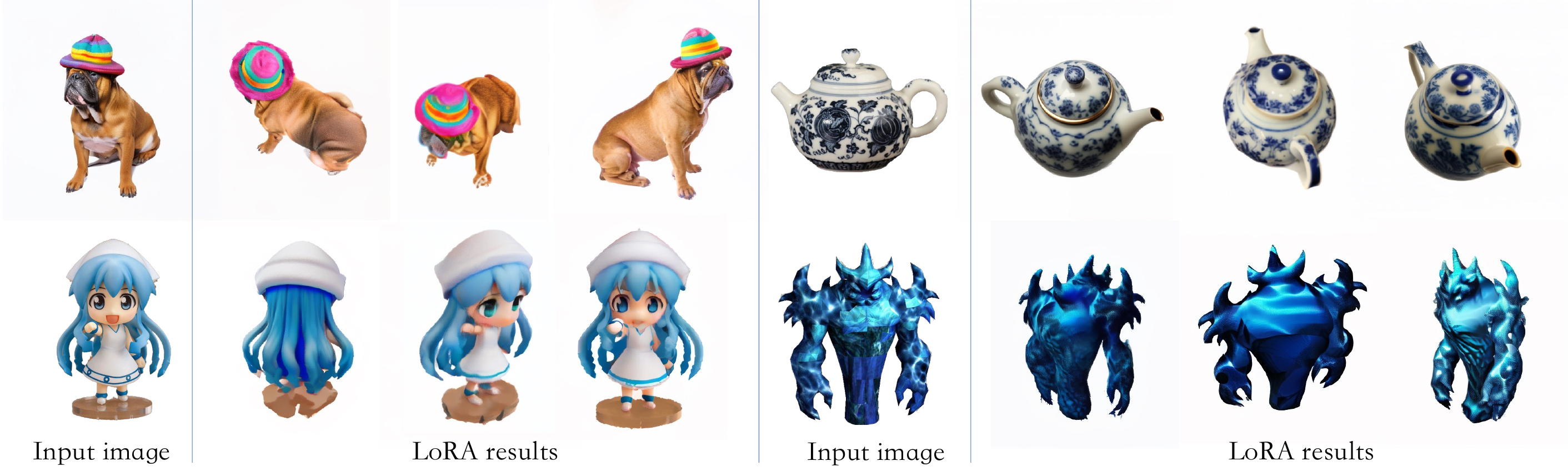}
   \caption{Results from LoRA after stage 2. Different images are obtained using different camera parameters as class embeddings and using no-texture rendering as base image.}
   \label{fig_lora}
\end{figure*}


\subsection{Stage2: ReFine NeRF}
\label{sec:s2}

In this stage, we continued to optimize based on the coarse NeRF.
After the first stage, we get a coarse NeRF and a pre-trained LoRA. We used the pre-trained LoRA to initialize LoRA in this stage. The training process is also divided into original view training and new view training. 

The original view training part is consistent with the stage 1, Eq.\ref{eq:ori} is used as the loss function for optimization.

In the new view training, we obtain the image $I$ and normal map $N_n$ through NeRF rendering. We use the noisy latent of the image $I$ as the input of the LoRA model and the original pre-trained T2I model to obtain the corresponding view results respectively, and calculate the Variational Score Distillation loss using Eq.\ref{eq:vsd}.

In this stage, LoRA is still trained through the images by NeRF rendering using Eq.\ref{eq:lora}. Different from stage 1, the noise level sampling range used by the LoRA model needs to be reduced as shown in Fig.\ref{fig2}. Therefore, in this stage, the loss function $L_{s2}$ we use to optimization is:

\begin{equation}
L_{p2}= \lambda_{vsd} L_{VSD} + \lambda_{3d}L_{SDS}^{3D}
  \label{eq:p2}
\end{equation}

\begin{equation}
L_{s2}= L_{ori} + L_{p2} + L_{normal} + L_{lora}
  \label{eq:S2}
\end{equation}

The reason for training LoRA in advance in stage 1 is to make LoRA conform to the current object as much as possible. In the original VSD\cite{wang2023prolificdreamer}, only using prompt to sample 3D information from the T2I model makes it difficult to control the details of 3D generation. On the other hand, it will cause the model generated in the image-to-3D task to be too different from the original image. Therefore, we pre-train LoRA using object-level rendering data in stage 1 and control the optimization range of LoRA from prompt-level to object-level. After stage 2 training, LoRA will be able to generate multi-view image corresponding to the input image using image-to-image method as shown in Fig.\ref{fig_lora}, which shows that the trained LoRA already has object-level 3D prior.

\subsection{Stage3: Refine 3D model}
\label{sec:s3}

After stage 2, we get a refined NeRF and a object-level LoRA model. NeRF can render high-quality image results, but the extracted mesh is coarse. In this stage, we will optimize the extracted mesh to achieve the same high-quality as NeRF rendering.

When extracting a model from NeRF, we usually need to use a threshold to determine the position of the mesh extraction surface. After determining the vertices to extract the mesh, we can get the color of vertices through the vertices positions, and then we unwrap the UV coordinates of the mesh using Xatlas\cite{xatlas}. In this way, we get a 3D model with UVmap, mesh $M=\{ Vec,F,UV \}$. We will optimize the UV-corresponding to mesh vertices $Vec$ and $UV$, in order to obtain a high-quality mesh.

During the 3D mesh rendering process, the camera intrinsics are aligned with the stage 2 to ensure that images of same views as the previous two stages can be obtained. We assign a trainable offset $\Delta v_i$ to each vertex $v_i$, and assign a texture offset $\Delta UV$ to the UVmap. During the rendering process:

\begin{equation}
 I_{3d}^c = f(Vec+\Delta Vec, UV+MLP(\Delta UV'), F, c)
  \label{eq:render}
\end{equation}

where $f$ represents the differentiable renderer, $F$ represents the faces of the mesh, $c$ represents the camera extrinsics of rendering and $MLP$ represents a multi-layer perceptron, which will calculate the real $\Delta UV$. When using $\Delta UV$ directly without using $MLP$ for mapping, the optimization effect is not ideal. During the optimization process, we will also divide it into the original view and the new view. The original view uses the original image to calculate the loss like Eq.\ref{eq:ori}. In the new view, we use the LoRA model trained in previous two stages as our pre-trained model to optimize the parameters, the gradient of rendering image $I_{3d}^c$ can be computed as follow:

\begin{equation}
\nabla L_{I3d} =\mathbb{E}_{t,\epsilon,c}\left[w_t\left(\epsilon_{lora}\left(I_{3d}^c;t,y,c\right)-\epsilon\right)\right]
  \label{eq:sds_mesh}
\end{equation}

The LoRA model is able to generate an image with better similarity to our current object than the original T2I model.

In this stage, the LoRA model is no longer trained. To prevent abrupt geometry, we apply a normal smoothing loss Eq.\ref{eq:normal} on the rendering image and add an L2 loss to $\Delta v_i$.

\begin{equation}
 L_{offset} = \sum_i(\Delta v_i)^2
  \label{eq:offset}
\end{equation}

These loss will prevent our vertex optimization from being too far away from the original position while ensuring the smoothness of the mesh.


\begin{figure*}[ht]
  \centering
  \includegraphics[scale=0.48]{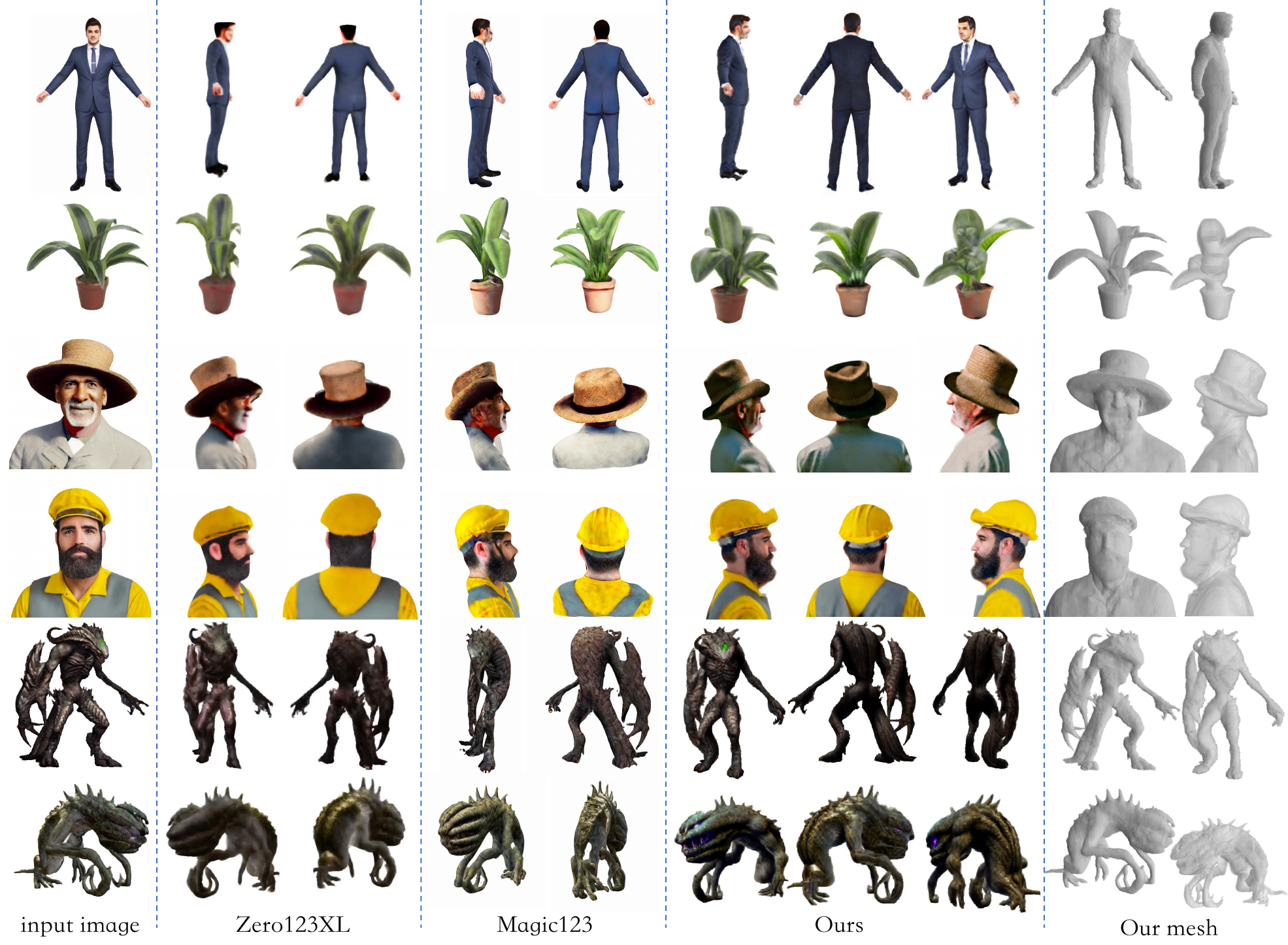}
  \caption{Qualitative comparisons of different methods. Compared with Magic123\cite{qian2023magic123} and Zero123XL\cite{liu2023zero}, our method performs better on both texture and 3D structure. The last column is the no-texture rendering results of the mesh obtained by our pipeline.}
  \label{figexp1}
\end{figure*}
\begin{figure*}[ht]
  \centering
  \includegraphics[scale=0.53]{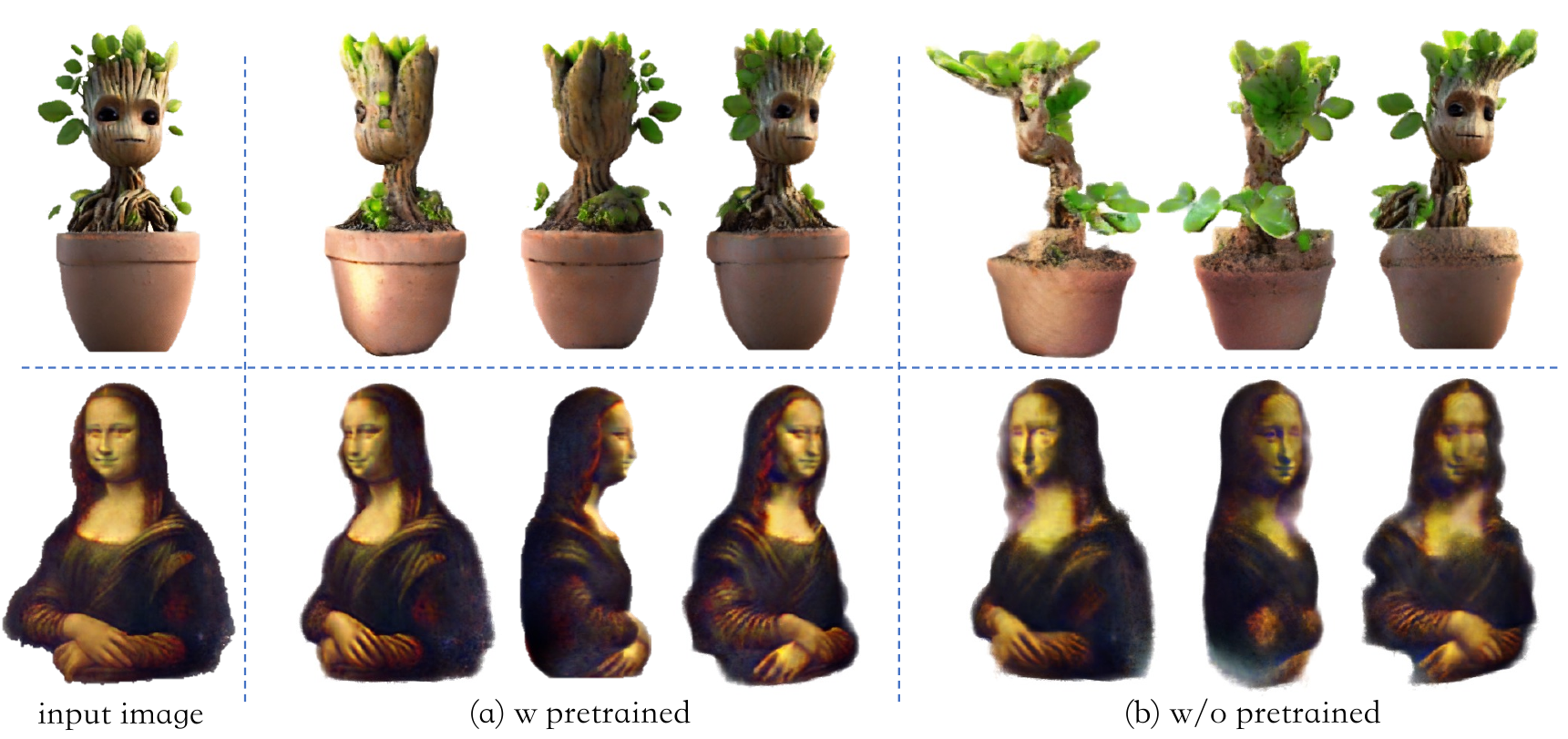}
  \caption{Ablation study of LoRA pre-trained in stage 1. (a) Trained by our presented method. (b) Without the pretraining of LoRA in stage 1.}
  \label{figexp2}
\end{figure*}

\section{Experiments}

\subsection{Implementation Details}

In all experiments, the basic model and optimizer used by all methods are same. We adopt the stable diffusion\cite{ho2020diffusion} v2.1-base version as pre-trained text-to-image model, and Zero123XL\cite{liu2023zero} as 3d prior diffusion model. We use Blip2\cite{li2023blip2} to generate the prompt corresponding to the input image. During the training phase, Adam is used for optimization, and the learning rate is set to 0.0001.
We use multi-scale hash encoding in Instant-NGP \cite{muller2022instant} as the basic model for NeRF in stages 1 and 2,
and use pytorch3d\cite{pytorch3d} as differentiable renderer in stage 3.In stage 1, we trained 1500 steps. The rendering resolution was set to 64 in the first 500 steps and 128 in the last 1000 steps. In stage 2, the resolution of novel view is set to 256, the resolution of original view is set to 512, and 3500 steps are trained. In stage 3, the resolution is set to 800 for mesh optimization, and trained 2000 steps. At stage 3, the mesh is extracted at a resolution of $512^3$ with a density threshold of 10 by marching cubes from NeRF trained in stage 2.

${\lambda}_{SDS}$ and ${\lambda}_{3d}$ are set to 0.2 and 1 for stage 1 and $\lambda_{vsd}$ is set to 1 in stage 2, which reduces the oversaturation of the texture. The loss weights ${\lambda}_{rgb}$ for color are linearly increased from 100 to 1000 during training, ${\lambda}_{mask}$ linearly increased from 50 to 500 during training, and the ${\lambda}_{normal}$ is increased from 0 to 100 in the first two stages and reduced from 100 to 10 in stage 3. In the training process of NeRF, we use pure white as the background.

In the training process, we assume that the input image is shot from the front view, that is, the initial polar angle is 90° and the azimuth angle is 0°. During the new view training, we will randomly sample the azimuth angle within 360° and the camera polar angle between 60 and 150, but keep the distance from the camera to the center of the object unchanged throughout the training process. At the same time, the intrinsics parameters of camera are all fixed during the training process. In the training process, it is only necessary to ensure that the rendering range of NeRF is within the range of the camera, and the intrinsics parameters of camera does not need to use a specific value.

\begin{figure}
	\centering
	\subfloat[Input image]{\includegraphics[width=.45\columnwidth]{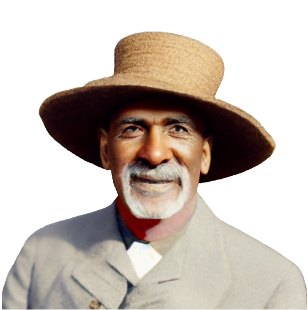}}\hspace{5pt}
	\subfloat[Mesh]{\includegraphics[width=.45\columnwidth]{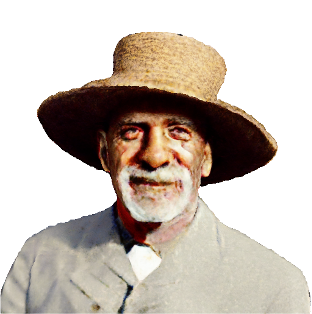}}\\
	\subfloat[DMtet]{\includegraphics[width=.45\columnwidth]{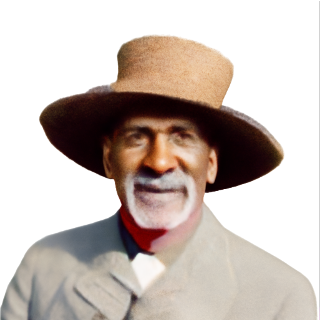}}\hspace{5pt}
	\subfloat[Stage3]{\includegraphics[width=.45\columnwidth]{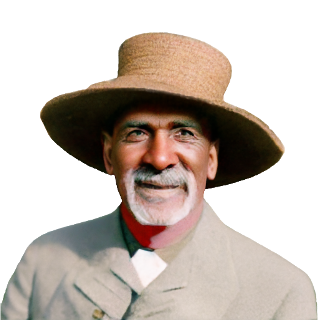}}
	\caption{Effect of stage 3. The rendering result (d) using our stage 3 refinement strategy is more consistent with the original input image than using DMtet (c).}
        \label{expab2}
\end{figure}

\subsection{Results and Comparisons}

\textbf{Qualitative Comparisons.} Our method will be compared with the state-of-the-art Zero123XL and Magic123. For Zero123XL, we use 3D-SDS loss to optimize a NeRF with the same parameters as our method. For Magic123, we use the original code, but replace the pre-trained diffusion model from v1.5 to v2.1-base version, and replace 3d prior from Zero123 to Zero123XL with higher performance, which yields better quality than the original implementation.

In Fig.\ref{figexp1}, we show the comparison results of our method with Zero123XL and Magic123. Our method achieves the best effect in texture performance and 3D structure. It is worth noting that our method can still generate very reasonable structure and fine texture in the case of rare objects, such as the monster related images in the last two lines, which also shows the robustness of our method.

\noindent
\textbf{Quantitative Evaluation.} We used the indicators employed in previous studies\cite{qian2023magic123}: PSNR and CLIP-Similarity\cite{clippaper}. We used a self-built dataset for evaluation, which contains real images similar to the input image shown in Fig.\ref{figexp1}. PSNR is measured in the original view of results to measure the reconstruction quality. Clip-similarity calculates the average clip distance between the rendered image and the input image, and measures the 3D consistency through the appearance similarity between the new view and the original view.

As shown in Table.\ref{table1}, compared with previous methods, our method achieves first-class performance in all metrics. Among them, ZeroXL-DMtet represents the result of refinement using DMtet, Ours-DMtet represents the result of optimization using DMtet and original diffusion in stage 3, and Ours-mesh represents the result of the final mesh of our method. PSNR results show that our method can restore the input better than other methods. The improvement of CLIP-Similarity reflects that our results have better 3D consistency.

\begin{table}
\begin{center}
\caption{We show the quantitative results based on CLIP-Similarity/PSNR. The bold is the best. ZeroXL-DMtet represents the result of refinement using DMtet, Ours-DMtet represents the result of optimization using DMtet and original diffusion in stage 3, and Ours-mesh represents the result of the final mesh of our method. 
}
\label{table1}
\begin{tabular}{c|c|c} 
\hline   Algorithms & PSNR↑ & CLIP-Similarity↑ \\   
\hline    
   Magic123 & 25.56 & 0.74  \\  
   Zero123XL & 21.32 & 0.61  \\ 
   Zero123XL-DMtet& 22.91 & 0.69  \\ 
\hline   Ours & \textbf{26.13} & 0.78  \\  
Ours-DMtet& 22.15 & 0.75  \\ 
Ours-mesh& 24.45 & \textbf{0.81}  \\ 
\hline
\end{tabular}
\end{center}
\end{table}

\subsection{Ablation Study}

\textbf{The effect of pre-training LoRA in the stage 1.} In Fig.\ref{figexp2}, we study the impact of LoRA training process on the results. It is obvious that without the pre-train of LoRA, the directly combination of VSD loss and 3D SDS will not generate a reasonable structure, and there may be a multi-face effect, as shown in the last line. Therefore, in our method, LoRA pre training in stage 1 is a necessary process.

\noindent
\textbf{Effect of stage 3}. We show the effect of stage 3 on the results in Fig.\ref{expab2}. (a) is the input image, (b) is the mesh extracted from the trained NeRF, and (c) is the effect of using Deep Marching Tetrahedra (DMTet)~\cite{dmtet} and original SDS loss to replace stage 3. It can be seen that the texture of (c) is relatively fuzzy. The texture generated after stage 3 (d) is more detailed, and the rendering result will be more consistent with the original input image. It can be observed an intuitive improvement in the quality of the final mesh using the proposed method.

\subsection{Limitations}

Although our method can achieve precise and robust 3D content generation, the overall time consumption is relatively high, requiring about than an hour of training time. We will optimize the speed using faster 3D representation in future work.

\section{Conclusion}

We have presented Boosting3D, a multi-stage pipeline, for 3D generation tasks guided by a single image. Benefiting from the boosted 3D prior (object-specific LoRA) , Boosting3D can produce reasonably fine results in different data domains and has high robustness for zero-shot images. Boosting3D outperforms previous technologies in terms of structural rationality and texture details, as demonstrated by experiments based on real and synthetic images. By optimizing the mesh, Boosting3D can obtain high-precision mesh results with high-quality texture representation. We believe that this work can effectively promote the development of universal 3D generation and has great potential in future applications.

\clearpage
{
    \small
    \bibliographystyle{ieeenat_fullname}
    \bibliography{main}

\begin{thebibliography}{43}
\providecommand{\natexlab}[1]{#1}
\providecommand{\url}[1]{\texttt{#1}}
\expandafter\ifx\csname urlstyle\endcsname\relax
  \providecommand{\doi}[1]{doi: #1}\else
  \providecommand{\doi}{doi: \begingroup \urlstyle{rm}\Url}\fi

\bibitem[Cao et~al.(2023)Cao, Kreis, Fidler, Sharp, and Yin]{cao2023texfusion}
Tianshi Cao, Karsten Kreis, Sanja Fidler, Nicholas Sharp, and Kangxue Yin.
\newblock Texfusion: Synthesizing 3d textures with text-guided image diffusion
  models.
\newblock In \emph{Proceedings of the IEEE/CVF International Conference on
  Computer Vision}, pages 4169--4181, 2023.

\bibitem[Chen et~al.(2023)Chen, Chen, Jiao, and Jia]{chen2023fantasia3d}
Rui Chen, Yongwei Chen, Ningxin Jiao, and Kui Jia.
\newblock Fantasia3d: Disentangling geometry and appearance for high-quality
  text-to-3d content creation.
\newblock \emph{arXiv preprint arXiv:2303.13873}, 2023.

\bibitem[Deng et~al.(2009)Deng, Dong, Socher, Li, Li, and
  Fei-Fei]{deng2009imagenet}
Jia Deng, Wei Dong, Richard Socher, Li-Jia Li, Kai Li, and Li Fei-Fei.
\newblock Imagenet: A large-scale hierarchical image database.
\newblock In \emph{2009 IEEE conference on computer vision and pattern
  recognition}, pages 248--255. Ieee, 2009.

\bibitem[Duggal and Pathak(2022)]{duggal2022topologically}
Shivam Duggal and Deepak Pathak.
\newblock Topologically-aware deformation fields for single-view 3d
  reconstruction.
\newblock In \emph{Proceedings of the IEEE/CVF Conference on Computer Vision
  and Pattern Recognition}, pages 1536--1546, 2022.

\bibitem[Gao et~al.(2022)Gao, Shen, Wang, Chen, Yin, Li, Litany, Gojcic, and
  Fidler]{gao2022get3d}
Jun Gao, Tianchang Shen, Zian Wang, Wenzheng Chen, Kangxue Yin, Daiqing Li, Or
  Litany, Zan Gojcic, and Sanja Fidler.
\newblock Get3d: A generative model of high quality 3d textured shapes learned
  from images.
\newblock \emph{Advances In Neural Information Processing Systems},
  35:\penalty0 31841--31854, 2022.

\bibitem[Guo et~al.(2023)Guo, Liu, Wang, Zou, Luo, Chen, Cao, and
  Zhang]{guo2023threestudio}
Yuan-Chen Guo, Ying-Tian Liu, Chen Wang, Zi-Xin Zou, Guan Luo, Chia-Hao Chen,
  Yan-Pei Cao, and Song-Hai Zhang.
\newblock threestudio: A unified framework for 3d content generation.
\newblock \url{https://github.com/threestudio-project/threestudio}, 2023.

\bibitem[Ho et~al.(2020)Ho, Jain, and Abbeel]{ho2020diffusion}
Jonathan Ho, Ajay Jain, and Pieter Abbeel.
\newblock Denoising diffusion probabilistic models.
\newblock \emph{Advances in neural information processing systems},
  33:\penalty0 6840--6851, 2020.

\bibitem[Hu et~al.(2021)Hu, Shen, Wallis, Allen-Zhu, Li, Wang, Wang, and
  Chen]{hu2021lora}
Edward~J Hu, Yelong Shen, Phillip Wallis, Zeyuan Allen-Zhu, Yuanzhi Li, Shean
  Wang, Lu Wang, and Weizhu Chen.
\newblock Lora: Low-rank adaptation of large language models.
\newblock \emph{arXiv preprint arXiv:2106.09685}, 2021.

\bibitem[Huang et~al.(2023)Huang, Wang, Shi, Qi, Zha, and
  Zhang]{huang2023dreamtime}
Yukun Huang, Jianan Wang, Yukai Shi, Xianbiao Qi, Zheng-Jun Zha, and Lei Zhang.
\newblock Dreamtime: An improved optimization strategy for text-to-3d content
  creation.
\newblock \emph{arXiv preprint arXiv:2306.12422}, 2023.

\bibitem[Jun and Nichol(2023)]{jun2023shape}
Heewoo Jun and Alex Nichol.
\newblock Shap-e: Generating conditional 3d implicit functions.
\newblock \emph{arXiv preprint arXiv:2305.02463}, 2023.

\bibitem[Kolotouros et~al.(2019)Kolotouros, Pavlakos, and
  Daniilidis]{kolotouros2019convolutional}
Nikos Kolotouros, Georgios Pavlakos, and Kostas Daniilidis.
\newblock Convolutional mesh regression for single-image human shape
  reconstruction.
\newblock In \emph{Proceedings of the IEEE/CVF Conference on Computer Vision
  and Pattern Recognition}, pages 4501--4510, 2019.

\bibitem[Li et~al.(2023)Li, Li, Savarese, and Hoi]{li2023blip2}
Junnan Li, Dongxu Li, Silvio Savarese, and Steven Hoi.
\newblock Blip-2: Bootstrapping language-image pre-training with frozen image
  encoders and large language models.
\newblock \emph{arXiv preprint arXiv:2301.12597}, 2023.

\bibitem[Lin et~al.(2023)Lin, Gao, Tang, Takikawa, Zeng, Huang, Kreis, Fidler,
  Liu, and Lin]{lin2023magic3d}
Chen-Hsuan Lin, Jun Gao, Luming Tang, Towaki Takikawa, Xiaohui Zeng, Xun Huang,
  Karsten Kreis, Sanja Fidler, Ming-Yu Liu, and Tsung-Yi Lin.
\newblock Magic3d: High-resolution text-to-3d content creation.
\newblock In \emph{Proceedings of the IEEE/CVF Conference on Computer Vision
  and Pattern Recognition}, pages 300--309, 2023.

\bibitem[Liu et~al.(2023{\natexlab{a}})Liu, Xu, Jin, Chen, Xu, Su,
  et~al.]{liu2023one}
Minghua Liu, Chao Xu, Haian Jin, Linghao Chen, Zexiang Xu, Hao Su, et~al.
\newblock One-2-3-45: Any single image to 3d mesh in 45 seconds without
  per-shape optimization.
\newblock \emph{arXiv preprint arXiv:2306.16928}, 2023{\natexlab{a}}.

\bibitem[Liu et~al.(2023{\natexlab{b}})Liu, Wu, Van~Hoorick, Tokmakov,
  Zakharov, and Vondrick]{liu2023zero}
Ruoshi Liu, Rundi Wu, Basile Van~Hoorick, Pavel Tokmakov, Sergey Zakharov, and
  Carl Vondrick.
\newblock Zero-1-to-3: Zero-shot one image to 3d object.
\newblock In \emph{Proceedings of the IEEE/CVF International Conference on
  Computer Vision}, pages 9298--9309, 2023{\natexlab{b}}.

\bibitem[Liu et~al.(2023{\natexlab{c}})Liu, Lin, Zeng, Long, Liu, Komura, and
  Wang]{liu2023syncdreamer}
Yuan Liu, Cheng Lin, Zijiao Zeng, Xiaoxiao Long, Lingjie Liu, Taku Komura, and
  Wenping Wang.
\newblock Syncdreamer: Generating multiview-consistent images from a
  single-view image.
\newblock \emph{arXiv preprint arXiv:2309.03453}, 2023{\natexlab{c}}.

\bibitem[Melas-Kyriazi et~al.(2023)Melas-Kyriazi, Laina, Rupprecht, and
  Vedaldi]{melas2023realfusion}
Luke Melas-Kyriazi, Iro Laina, Christian Rupprecht, and Andrea Vedaldi.
\newblock Realfusion: 360deg reconstruction of any object from a single image.
\newblock In \emph{Proceedings of the IEEE/CVF Conference on Computer Vision
  and Pattern Recognition}, pages 8446--8455, 2023.

\bibitem[Metzer et~al.(2023)Metzer, Richardson, Patashnik, Giryes, and
  Cohen-Or]{metzer2023latent}
Gal Metzer, Elad Richardson, Or Patashnik, Raja Giryes, and Daniel Cohen-Or.
\newblock Latent-nerf for shape-guided generation of 3d shapes and textures.
\newblock In \emph{Proceedings of the IEEE/CVF Conference on Computer Vision
  and Pattern Recognition}, pages 12663--12673, 2023.

\bibitem[Mildenhall et~al.(2021)Mildenhall, Srinivasan, Tancik, Barron,
  Ramamoorthi, and Ng]{nerf}
Ben Mildenhall, Pratul~P Srinivasan, Matthew Tancik, Jonathan~T Barron, Ravi
  Ramamoorthi, and Ren Ng.
\newblock Nerf: Representing scenes as neural radiance fields for view
  synthesis.
\newblock \emph{Communications of the ACM}, 65\penalty0 (1):\penalty0 99--106,
  2021.

\bibitem[M{\"u}ller et~al.(2022)M{\"u}ller, Evans, Schied, and
  Keller]{muller2022instant}
Thomas M{\"u}ller, Alex Evans, Christoph Schied, and Alexander Keller.
\newblock Instant neural graphics primitives with a multiresolution hash
  encoding.
\newblock \emph{ACM Transactions on Graphics (ToG)}, 41\penalty0 (4):\penalty0
  1--15, 2022.

\bibitem[Poole et~al.(2022)Poole, Jain, Barron, and
  Mildenhall]{poole2022dreamfusion}
Ben Poole, Ajay Jain, Jonathan~T Barron, and Ben Mildenhall.
\newblock Dreamfusion: Text-to-3d using 2d diffusion.
\newblock \emph{arXiv preprint arXiv:2209.14988}, 2022.

\bibitem[Qian et~al.(2023)Qian, Mai, Hamdi, Ren, Siarohin, Li, Lee,
  Skorokhodov, Wonka, Tulyakov, et~al.]{qian2023magic123}
Guocheng Qian, Jinjie Mai, Abdullah Hamdi, Jian Ren, Aliaksandr Siarohin, Bing
  Li, Hsin-Ying Lee, Ivan Skorokhodov, Peter Wonka, Sergey Tulyakov, et~al.
\newblock Magic123: One image to high-quality 3d object generation using both
  2d and 3d diffusion priors.
\newblock \emph{arXiv preprint arXiv:2306.17843}, 2023.

\bibitem[Radford et~al.(2021)Radford, Kim, Hallacy, Ramesh, Goh, Agarwal,
  Sastry, Askell, Mishkin, Clark, et~al.]{clippaper}
Alec Radford, Jong~Wook Kim, Chris Hallacy, Aditya Ramesh, Gabriel Goh,
  Sandhini Agarwal, Girish Sastry, Amanda Askell, Pamela Mishkin, Jack Clark,
  et~al.
\newblock Learning transferable visual models from natural language
  supervision.
\newblock In \emph{International conference on machine learning}, pages
  8748--8763. PMLR, 2021.

\bibitem[Raj et~al.(2023)Raj, Kaza, Poole, Niemeyer, Ruiz, Mildenhall, Zada,
  Aberman, Rubinstein, Barron, et~al.]{raj2023dreambooth3d}
Amit Raj, Srinivas Kaza, Ben Poole, Michael Niemeyer, Nataniel Ruiz, Ben
  Mildenhall, Shiran Zada, Kfir Aberman, Michael Rubinstein, Jonathan Barron,
  et~al.
\newblock Dreambooth3d: Subject-driven text-to-3d generation.
\newblock \emph{arXiv preprint arXiv:2303.13508}, 2023.

\bibitem[Ravi et~al.(2020)Ravi, Reizenstein, Novotny, Gordon, Lo, Johnson, and
  Gkioxari]{pytorch3d}
Nikhila Ravi, Jeremy Reizenstein, David Novotny, Taylor Gordon, Wan-Yen Lo,
  Justin Johnson, and Georgia Gkioxari.
\newblock Accelerating 3d deep learning with pytorch3d.
\newblock \emph{arXiv preprint arXiv:2007.08501}, 2020.

\bibitem[Richardson et~al.(2023)Richardson, Metzer, Alaluf, Giryes, and
  Cohen-Or]{richardson2023texture}
Elad Richardson, Gal Metzer, Yuval Alaluf, Raja Giryes, and Daniel Cohen-Or.
\newblock Texture: Text-guided texturing of 3d shapes.
\newblock \emph{arXiv preprint arXiv:2302.01721}, 2023.

\bibitem[Ruiz et~al.(2023)Ruiz, Li, Jampani, Pritch, Rubinstein, and
  Aberman]{dreambooth}
Nataniel Ruiz, Yuanzhen Li, Varun Jampani, Yael Pritch, Michael Rubinstein, and
  Kfir Aberman.
\newblock Dreambooth: Fine tuning text-to-image diffusion models for
  subject-driven generation.
\newblock In \emph{Proceedings of the IEEE/CVF Conference on Computer Vision
  and Pattern Recognition}, pages 22500--22510, 2023.

\bibitem[Saito et~al.(2019)Saito, Huang, Natsume, Morishima, Kanazawa, and
  Li]{saito2019pifu}
Shunsuke Saito, Zeng Huang, Ryota Natsume, Shigeo Morishima, Angjoo Kanazawa,
  and Hao Li.
\newblock Pifu: Pixel-aligned implicit function for high-resolution clothed
  human digitization.
\newblock In \emph{Proceedings of the IEEE/CVF international conference on
  computer vision}, pages 2304--2314, 2019.

\bibitem[Saito et~al.(2020)Saito, Simon, Saragih, and Joo]{saito2020pifuhd}
Shunsuke Saito, Tomas Simon, Jason Saragih, and Hanbyul Joo.
\newblock Pifuhd: Multi-level pixel-aligned implicit function for
  high-resolution 3d human digitization.
\newblock In \emph{Proceedings of the IEEE/CVF Conference on Computer Vision
  and Pattern Recognition}, pages 84--93, 2020.

\bibitem[Sargent et~al.(2023)Sargent, Koh, Zhang, Chang, Herrmann, Srinivasan,
  Wu, and Sun]{sargent2023vq3d}
Kyle Sargent, Jing~Yu Koh, Han Zhang, Huiwen Chang, Charles Herrmann, Pratul
  Srinivasan, Jiajun Wu, and Deqing Sun.
\newblock Vq3d: Learning a 3d-aware generative model on imagenet.
\newblock \emph{arXiv preprint arXiv:2302.06833}, 2023.

\bibitem[Schuhmann et~al.(2022)Schuhmann, Beaumont, Vencu, Gordon, Wightman,
  Cherti, Coombes, Katta, Mullis, Wortsman, et~al.]{schuhmann2022laion}
Christoph Schuhmann, Romain Beaumont, Richard Vencu, Cade Gordon, Ross
  Wightman, Mehdi Cherti, Theo Coombes, Aarush Katta, Clayton Mullis, Mitchell
  Wortsman, et~al.
\newblock Laion-5b: An open large-scale dataset for training next generation
  image-text models.
\newblock \emph{Advances in Neural Information Processing Systems},
  35:\penalty0 25278--25294, 2022.

\bibitem[Shen et~al.(2021)Shen, Gao, Yin, Liu, and Fidler]{dmtet}
Tianchang Shen, Jun Gao, Kangxue Yin, Ming-Yu Liu, and Sanja Fidler.
\newblock Deep marching tetrahedra: a hybrid representation for high-resolution
  3d shape synthesis.
\newblock \emph{Advances in Neural Information Processing Systems},
  34:\penalty0 6087--6101, 2021.

\bibitem[Skorokhodov et~al.(2023)Skorokhodov, Siarohin, Xu, Ren, Lee, Wonka,
  and Tulyakov]{skorokhodov20233d}
Ivan Skorokhodov, Aliaksandr Siarohin, Yinghao Xu, Jian Ren, Hsin-Ying Lee,
  Peter Wonka, and Sergey Tulyakov.
\newblock 3d generation on imagenet.
\newblock \emph{arXiv preprint arXiv:2303.01416}, 2023.

\bibitem[Tang et~al.(2023{\natexlab{a}})Tang, Ren, Zhou, Liu, and
  Zeng]{dreamgaussian}
Jiaxiang Tang, Jiawei Ren, Hang Zhou, Ziwei Liu, and Gang Zeng.
\newblock Dreamgaussian: Generative gaussian splatting for efficient 3d content
  creation.
\newblock \emph{arXiv preprint arXiv:2309.16653}, 2023{\natexlab{a}}.

\bibitem[Tang et~al.(2023{\natexlab{b}})Tang, Wang, Zhang, Zhang, Yi, Ma, and
  Chen]{tang2023make}
Junshu Tang, Tengfei Wang, Bo Zhang, Ting Zhang, Ran Yi, Lizhuang Ma, and Dong
  Chen.
\newblock Make-it-3d: High-fidelity 3d creation from a single image with
  diffusion prior.
\newblock \emph{arXiv preprint arXiv:2303.14184}, 2023{\natexlab{b}}.

\bibitem[Tsalicoglou et~al.(2023)Tsalicoglou, Manhardt, Tonioni, Niemeyer, and
  Tombari]{ts2023textmesh}
Christina Tsalicoglou, Fabian Manhardt, Alessio Tonioni, Michael Niemeyer, and
  Federico Tombari.
\newblock Textmesh: Generation of realistic 3d meshes from text prompts.
\newblock \emph{arXiv preprint arXiv:2304.12439}, 2023.

\bibitem[Wang et~al.(2023)Wang, Lu, Wang, Bao, Li, Su, and
  Zhu]{wang2023prolificdreamer}
Zhengyi Wang, Cheng Lu, Yikai Wang, Fan Bao, Chongxuan Li, Hang Su, and Jun
  Zhu.
\newblock Prolificdreamer: High-fidelity and diverse text-to-3d generation with
  variational score distillation.
\newblock \emph{arXiv preprint arXiv:2305.16213}, 2023.

\bibitem[Xiu et~al.(2023)Xiu, Yang, Cao, Tzionas, and Black]{xiu2023econ}
Yuliang Xiu, Jinlong Yang, Xu Cao, Dimitrios Tzionas, and Michael~J Black.
\newblock Econ: Explicit clothed humans optimized via normal integration.
\newblock In \emph{Proceedings of the IEEE/CVF Conference on Computer Vision
  and Pattern Recognition}, pages 512--523, 2023.

\bibitem[Xu et~al.(2023)Xu, Jiang, Wang, Fan, Wang, and Wang]{xu2023neurallift}
Dejia Xu, Yifan Jiang, Peihao Wang, Zhiwen Fan, Yi Wang, and Zhangyang Wang.
\newblock Neurallift-360: Lifting an in-the-wild 2d photo to a 3d object with
  360deg views.
\newblock In \emph{Proceedings of the IEEE/CVF Conference on Computer Vision
  and Pattern Recognition}, pages 4479--4489, 2023.

\bibitem[Young(2021)]{xatlas}
Jonathan Young.
\newblock xatlas: Mesh parameterization / uv unwrapping library.
\newblock \url{https://github.com/jpcy/xatlas}, 2021.

\bibitem[Yu et~al.(2021)Yu, Ye, Tancik, and Kanazawa]{yu2021pixelnerf}
Alex Yu, Vickie Ye, Matthew Tancik, and Angjoo Kanazawa.
\newblock pixelnerf: Neural radiance fields from one or few images.
\newblock In \emph{Proceedings of the IEEE/CVF Conference on Computer Vision
  and Pattern Recognition}, pages 4578--4587, 2021.

\bibitem[Zheng et~al.(2019)Zheng, Yu, Wei, Dai, and Liu]{zheng2019deephuman}
Zerong Zheng, Tao Yu, Yixuan Wei, Qionghai Dai, and Yebin Liu.
\newblock Deephuman: 3d human reconstruction from a single image.
\newblock In \emph{Proceedings of the IEEE/CVF International Conference on
  Computer Vision}, pages 7739--7749, 2019.

\bibitem[Zhuang et~al.(2023)Zhuang, Wang, Liu, Lin, and
  Li]{zhuang2023dreameditor}
Jingyu Zhuang, Chen Wang, Lingjie Liu, Liang Lin, and Guanbin Li.
\newblock Dreameditor: Text-driven 3d scene editing with neural fields.
\newblock \emph{arXiv preprint arXiv:2306.13455}, 2023.

\end{thebibliography}
}


\end{document}